%% file: main.tex
\documentclass[conference]{IEEEtran}
\IEEEoverridecommandlockouts
\usepackage{cite}
\usepackage{amsmath,amssymb,amsfonts}

\usepackage{booktabs}
\usepackage{verbatim}
\usepackage{algorithmic}
\usepackage{graphicx}
\usepackage{textcomp}
\usepackage{xcolor}
\usepackage{url}
\usepackage{float}
\usepackage{svg}
\def\BibTeX{{\rm B\kern-.05em{\sc i\kern-.025em b}\kern-.08em
    T\kern-.1667em\lower.7ex\hbox{E}\kern-.125emX}}
\begin{document}

\title{The Dawn of KAN in Image-to-Image (I2I) Translation: Integrating Kolmogorov-Arnold Networks with GANs for Unpaired I2I Translation}
\author{
\IEEEauthorblockN{Arpan Mahara\IEEEauthorrefmark{1},
Naphtali D. Rishe\IEEEauthorrefmark{2}, Liangdong Deng\IEEEauthorrefmark{3}}
\IEEEauthorblockA{Knight Foundation School of Computing and Information Sciences\\
Florida International University\\
Miami, FL 33199\\
\IEEEauthorrefmark{1}amaha038@fiu.edu,
\IEEEauthorrefmark{2}rishen@cs.fiu.edu,
\IEEEauthorrefmark{2}liadeng@cs.fiu.edu}
}

\maketitle
\begin{abstract}
Image-to-Image translation in Generative Artificial Intelligence (Generative AI) has been a central focus of research, with applications spanning healthcare, remote sensing, physics, chemistry, photography, and more. Among the numerous methodologies, Generative Adversarial Networks (GANs) with contrastive learning have been particularly successful. This study aims to demonstrate that the Kolmogorov-Arnold Network (KAN) can effectively replace the Multi-layer Perceptron (MLP) method in generative AI, particularly in the subdomain of image-to-image translation, to achieve better generative quality. Our novel approach replaces the two-layer MLP with a two-layer KAN in the existing Contrastive Unpaired Image-to-Image Translation (CUT) model, developing the KAN-CUT model. This substitution favors the generation of more informative features in low-dimensional vector representations, which contrastive learning can utilize more effectively to produce high-quality images in the target domain. Extensive experiments, detailed in the results section, demonstrate the applicability of KAN in conjunction with contrastive learning and GANs in Generative AI, particularly for image-to-image translation. This work suggests that KAN could be a valuable component in the broader generative AI domain.
\end{abstract}

\begin{IEEEkeywords}
Generative AI, Image-to-Image translation, Generative Adversarial Networks (GANs), Contrastive Learning, Multi-layer Perceptron, Kolmogorov-Arnold Networks (KANs), PatchNCE Loss
\end{IEEEkeywords}

\section{Introduction}
Generative AI is prevalent across various research fields, including images, texts, videos, and more. It has been developed to achieve generative outcomes such as text-to-text \cite{vaswani2017attention}, text-to-image \cite{ramesh2021zero}, image-to-text \cite{xu2015show}, text-to-video \cite{wu2023tune}, video-to-video \cite{wang2018video}, and image-to-image generation or translation \cite{isola2017image}. The present study primarily focuses on image-to-image translation, a subdomain of Generative AI.

The popularity and success of Generative AI can be largely attributed to the flexibility and expressiveness of Multi-layer Perceptrons (MLPs) \cite{bishop1994neural, hornik1989multilayer, haykin1998neural}. Despite their significant contributions to the field of Deep Learning (a branch of Generative AI), MLPs have certain limitations, such as the inability to optimize univariate functions effectively and their lower accuracy compared to splines in low-dimensional spaces.

Recently, the Kolmogorov-Arnold Network (KAN) \cite{liu2024kan}, based on the Kolmogorov-Arnold representation theorem \cite{kolmogorov1961representation}, has been proposed as a potential replacement for MLPs. KAN combines the strengths of MLPs and splines, offering improved accuracy and interpretability. The authors have demonstrated KAN's better performance compared to MLPs in small-scale AI applications and suggested its potential in broader applications.

Image-to-image translation is a well-researched domain where images from domain $A$ are translated to domain $B$ using generative mechanisms to achieve various outcomes such as facial attribute manipulation \cite{hu2020facial}, medical image analysis \cite{song2024s}, and geospatial analysis \cite{mahara2024generative}. Among various mechanisms like VAEs \cite{zhao2021unpaired} and diffusion models \cite{zhao2022egsde}, GANs have been widely used to achieve image-to-image translation. Initially, the Pix2Pix model \cite{isola2017image} utilized paired images of domains $A$ and $B$ in a supervised manner for image-to-image translation. However, due to the impracticality of obtaining paired images, the CycleGAN model \cite{zhu2017unpaired} was introduced to perform translation in an unsupervised manner. Several other promising GANs equipped with unsupervised principles, such as GCGAN \cite{fu2019geometry}, CUT \cite{park2020contrastive}, DCLGAN \cite{han2021dual}, and StarGAN \cite{choi2018stargan}, have since been proposed. Among these, CUT is particularly notable for its accuracy, computational efficiency, and time complexity due to its uni-directional training. This model combines Generative Adversarial training with mutual information maximization using contrastive learning to generate high-quality images in the target domain.

A noteworthy aspect of this mechanism is its use of contrastive learning, inspired by the SIMCLR \cite{chen2020simple} framework, where features processed from different layers of the generator are fed into a two-layer MLP to enhance feature representation. This process allows for high-quality image generation in the target domain. 

Understanding the importance of KAN, to demonstrate and potentially prove the applicability of KAN in Generative AI, specifically in image-to-image translation, we first propose a novel customization of KAN and construct an efficient two-layer KAN, which we then use to replace the two-layer MLP in CUT, giving rise to the KAN-CUT model. Our overall contributions in the present study are:

\begin{itemize}
    \item Reformulated the KAN architecture to improve efficiency by avoiding the expansion of the input tensor and removing additional entropy regularization for L1 normalization.
    \item Enhanced the KAN layer by implementing an activation function that concatenates the basis function and spline function, replacing the original addition operation, and supplemented with additional processing using Gated Linear Units (GLU).
    \item Innovatively replaced the two-layer MLP in the CUT model with our efficient two-layer KAN, creating the KAN-CUT model for unpaired image-to-image translation.
    \item This study marks the first integration of KAN in the image-to-image translation domain (a subdomain of Generative AI), potentially paving the way for numerous applications of KAN in different Generative AI domains to achieve better outcomes.
\end{itemize}

\section{Related Work}
In unpaired image-to-image translation within the GANs domain, CycleGAN \cite{zhu2017unpaired} was a foundational work where images from domains $A$ and $B$ were trained on the cyclic principle. This principle dictates that if $b$ is an image generated by generator $G$ with $a$ as an input, then we should be able to obtain the original image $a$ when we feed the generated image $b$ to another generator $F$. The ultimate goal is to obtain a mapping of each instance from $A$ and $B$ in the absence of paired images, so that a previously unseen instance image from domain $A$ can be accurately translated to another instance of an image in domain $B$. While the model was very successful and is still being applied in various research endeavors, it has limitations such as being restrictive, computationally expensive, and having high time complexity due to its cyclic nature.

To address limitations related to time complexity and computational expense, several different works in GANs have been presented that make the training process single-directional, thus eliminating the need for auxiliary generators and discriminators. Notable examples include GCGAN \cite{fu2019geometry} and CUT \cite{park2020contrastive}. GCGAN leverages geometric consistency to impose the structural correspondence between the input and output images, ensuring that the generated image maintains the geometrical structure of the input image. 

Contrastive Learning, a methodology that favors achieving numerous Machine Learning (ML) and Deep Learning tasks such as classification, segmentation, simulation, and generation, operates with a self-supervised mechanism in the absence of supervised data. SIMCLR, a contrastive framework presented in \cite{chen2020simple}, was a remarkable work in Contrastive Learning based on image data that obtained better results than previous semi-supervised and self-supervised tasks, and comparable performance to well-known supervised models like ResNet50 \cite{he2016deep} in the ImageNet dataset. In their work, the authors showcased that simple data augmentation and applying contrastive learning to the patches of augmented images can help determine similar features in embedding space, and maximizing mutual information can lead to the final goal without the need for human-labeled data. More importantly, the authors of SIMCLR \cite{chen2020simple} found that instead of directly applying contrastive learning to features or vectors in lower dimensions, processing the features with a two-layer MLP before applying contrastive learning yielded better results.

Inspired by SIMCLR, \cite{park2020contrastive} proposed the CUT model, which utilized contrastive learning at the patch level, integrated with GAN in a uni-directional fashion, to address the limitations of CycleGAN. This model successfully achieved better performance based on quantitative evaluation along with lower time complexity. The most important procedure in this method is that during training, patches drawn from the input image and target domain image are processed through network layers and propagated to a two-layer MLP to obtain enriched features or vectors, following \cite{chen2020simple}. In the obtained enriched vectors, contrastive learning is carried out, and maximization of information among corresponding patches is done to maintain qualitative and selective image generation. It can be seen that the two-layer MLP is a crucial component that helps generate enriched vectors or features used for image-to-image translation.

Some subsequent works focused on improving the quality of generation while still incorporating bi-directional training with the need for two generators and two discriminators but with additional changes in the principle. Examples of these works include DualGAN \cite{yi2017dualgan} and DCLGAN \cite{han2021dual}. DualGAN introduced a dual learning mechanism inspired by natural language translation. It utilizes a primary GAN to convert images from the input domain to the output domain and a secondary GAN to reverse this conversion. This architecture ensures consistent and accurate mappings between domains by using reconstruction loss, which enhances the quality and robustness of the generated images. Similarly, DCLGAN \cite{han2021dual} shows that reverse mapping can be achieved without depending on the generated images, leading to non-restrictive mapping, unlike the mechanism of CycleGAN \cite{zhu2017unpaired}. It aligns with CycleGAN \cite{zhu2017unpaired} by using two generators and two discriminators and bi-directional training, additionally applying contrastive learning, similar to CUT \cite{park2020contrastive}, by maximizing mutual information among patches in both directions. DCLGAN can be considered an upgraded version of CUT. These GANs, successful in unpaired image-to-image translation, were mainly constructed to function between two domains. However, there are scenarios where generation is needed among multiple domains, such as generating images from various attributes or styles. StarGAN \cite{choi2018stargan} was proposed to achieve multi-domain translation using a single generator and discriminator. StarGAN \cite{choi2018stargan} was proposed to achieve translation across multiple domains using a unified generator and discriminator. StarGAN acquires mappings among various domains by conditioning the generator on domain labels, allowing it to flexibly translate an input image to any desired target domain.

\begin{figure*}[t]
  \centering
  \includegraphics[width=1\linewidth]{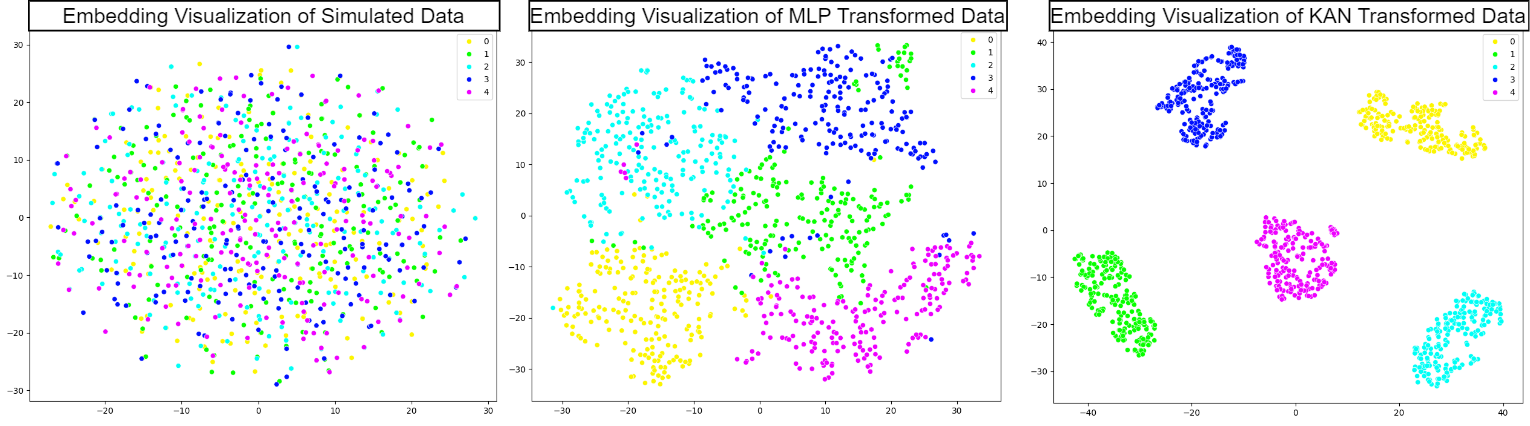}
  \caption{Comparison of Embedding Visualizations of Simulated Data with Transformation by Simple MLP and KAN Models}
  \label{fig:Feature_Representation}
\end{figure*}

KANs \cite{liu2024kan}, recently proposed, have been applied to various Machine Learning and Deep Learning tasks, such as image classification \cite{cheon2024kolmogorov}, image segmentation and generation \cite{li2024u}, time-series analysis \cite{vaca2024kolmogorov}, and graph-based learning \cite{kiamari2024gkan}. Even though \cite{li2024u} presents a generative model called Diffusion U-KAN, it only depicts generating images based on training limited to one domain. Due to the lack of work in the generation or translation of images from one domain to another, we propose to integrate KAN into the existing CUT model \cite{park2020contrastive} and present the new KAN-CUT model, capable of generating higher quality images across two domains. The code is available at \textbf{https://github.com/amaha7984/KAN-CUT}.

\section{Proposed Approach}
In this section, we first detail the architecture of the Kolmogorov-Arnold Network (KAN), followed by our novel changes and enhancements to the architecture. We then proceed to integrate the customized efficient KAN into the domain of image-to-image translation by replacing the two-layer Multi-layer Perceptron (MLP) with a two-layer KAN in the Contrastive Unpaired Image-to-Image Translation (CUT) \cite{park2020contrastive} model, resulting in the KAN-CUT model.

\subsection{Kolmogorov-Arnold Networks (KAN)}
Kolmogorov-Arnold Networks (KANs) have been discussed as a significant advancement in machine learning, often referred to as Machine Learning 2.0 among researchers. At a high level, our approach facilitates the generation of better-informed features in lower dimensions where contrastive learning can be performed. Due to the lack of previous research on KANs in representation learning, it is not straightforward to deduce their relevance for understanding or generating feature vectors in embedding space better than the well-established MLPs. To build our initial confidence, we conducted a simulation where data had three different labels (categories) with various features in a simulated embedding space. After training simple MLP and KAN models separately on the given simulated data, we visualized the results using t-SNE \cite{van2008visualizing} and observed that KANs performed better in clustering similar data points in the feature representation, as depicted in Fig. \ref{fig:Feature_Representation}. 

Before exploring how KAN can be used in generative tasks, we first discuss the fundamentals of KAN. Unlike MLP, which is based on the Universal Approximation Theorem, KAN is based on the Kolmogorov-Arnold Representation Theorem \cite{kolmogorov1961representation}. This theorem states that if $f$ is a multivariate continuous function on a bounded domain, then $f$ can be simplified into a finite composition of continuous single-variable functions and the binary operation of addition:

\begin{equation}
f(x) = f(x_1, \cdots, x_n) = \sum_{q=1}^{2n+1} \Phi_q \left( \sum_{p=1}^{n} \varphi_{q,p}(x_p) \right) \tag{2.1}
\end{equation}
where $\varphi_{q,p} : [0,1] \rightarrow \mathbb{R}$ and $\Phi_q : \mathbb{R} \rightarrow \mathbb{R}$.

Our main goal is to translate or generate images from one domain, say domain $A$, to another domain, say domain $B$. For this, we seek a function $g$ such that $g(A)$ yields $B' \approx B$. According to \cite{liu2024kan}, we can approximate $g$ by learning a discrete number of one-dimensional functions, parameterized by B-spline curves with tunable coefficients of local B-spline functions. 

KAN is comparable to MLP, but instead of learnable weights, it has learnable activation functions on edges, and summation on the resultant learned function's output is performed at the nodes. As explained in \cite{liu2024kan}, a KAN layer with $n_{\text{in}}$-dimensional inputs and $n_{\text{out}}$-dimensional outputs can be represented as a matrix of 1D functions:
\[
\Phi = \{\varphi_{q,p}\}, \quad p = 1,2, \cdots, n_{\text{in}}, \quad q = 1,2 \cdots, n_{\text{out}}, \tag{2.2}
\]
where each function's ($\varphi_{q,p}$) parameters are trainable. In the Kolmogorov-Arnold theorem, the internal functions constitute a KAN layer with $n_{\text{in}} = n$ and $n_{\text{out}} = 2n+1$, while the external functions form another KAN layer with $n_{\text{in}} = 2n+1$ and $n_{\text{out}} = 1$. Thus, the Kolmogorov-Arnold representations in Eq.~(2.1) are constructed by composing two KAN layers.

\subsection{B-Splines}
B-splines are piecewise polynomial curves constructed from a sequence of lower-order polynomial segments. They are defined by a set of control points \( P_i \) and a knot vector \( u \) that determines the influence of each control point on the B-spline curve.

The basis function for B-splines of order 0 (piecewise constant) is defined as:
\[
M_{i,0}(u) = 
\begin{cases} 
1 & \text{if } u_i \leq u < u_{i+1} \\
0 & \text{otherwise} 
\end{cases}
\]

For higher-order basis functions (piecewise linear, quadratic, etc.), the recursive definition is:
\[
M_{i,k}(u) = \frac{u - u_i}{u_{i+k} - u_i} M_{i,k-1}(u) + \frac{u_{i+k+1} - u}{u_{i+k+1} - u_{i+1}} M_{i+1,k-1}(u)
\]
where \( k = 1, \ldots, d \), and \( d \) is the degree of the B-spline.

The B-spline curve is then defined by:
\[
c(u) = \sum_{i=0}^{m} P_i M_{i,d}(u)
\]
where \( P_i \) are the control points, \( M_{i,d}(u) \) are the basis functions of degree \( d \), and \( u \) is the parameter.

In this context, \( u_i \) represents the elements of the knot vector, and \( m \) corresponds to the total number of control points minus one.

We have demonstrated the importance of the two-layer MLP in the image-to-image translation domain, particularly in applications involving contrastive learning and generative adversarial networks, as mentioned in Section II. We have also explored the concept of the KAN network and its potential applicability. Below, we will introduce our novel changes and refinements of the KAN network, which we will later apply to generative mechanisms.

\subsection{Efficient Two-Layer KAN}
For a layer with $\text{in\_features}$ inputs and $\text{out\_features}$ outputs, the original implementation expands the input to a tensor of the shape $(\text{batch\_size}, \text{out\_features}, \text{in\_features})$ to apply the activation functions. As mentioned above in Subsection A, all the activation functions are linear combinations of a fixed set of basis functions, B-spline curves. Following the work of \cite{efficientkan}, for efficient processing, we reconstructed the computation process by refraining from the additional step of expanding the input. Instead, we apply learnable activation functions directly to the input tensor and then combine the results linearly. Mathematically, if $\mathbf{i}$ is the input, and $\mathbf{B}(\mathbf{i})$ represents the B-spline basis functions applied to the input, the reformulated procedure of applying the activation function $\sigma$ on the input can be expressed as:

\[
\sigma(\mathbf{i}) = \sigma_b(\mathbf{i}) + \mathbf{B}(\mathbf{i})
\]

where $\sigma_b(\mathbf{i})$ is the base activation function applied to the input $\mathbf{i}$.

This modification simplifies the computation to a straightforward matrix multiplication, efficiently supporting both forward and backward passes. Similarly, regarding regularization, L1 regularization \cite{tibshirani1996regression} is applied in MLP to achieve sparsity and improve model interpretability. In the original KAN \cite{liu2024kan}, L1 regularization was adapted for learnable activation functions rather than linear weights since there are no learnable weights in terms of KAN. Specifically, the L1 norm was defined as the average magnitude of these activation functions over their inputs. This approach required the activation functions to be sparsified by their L1 norm. Additionally, an entropy regularization was necessary to further enforce sparsity. In our implementation of KAN, we align with the original approach by applying L1 regularization to the B-spline activation functions (`spline\_activation`) to enforce sparsity and control overfitting. The main difference is the removal of the tensor expansion step and additional entropy regularization, making the process more efficient.

To obtain an optimizable layer, the original KAN implementation \cite{liu2024kan} used a residual connection mechanism and expressed the activation function \(\phi(x)\) as the combination of the basis function \(b(x)\) and the spline function, given as:

\[ \phi(x) = w_b b(x) + w_s \text{spline}(x) \]

where $w_b$ and $w_s$ are learnable weights, but they can be considered redundant since they can be merged into \(b(x)\) and spline(x) \cite{liu2024kan}.

Instead of the additive operation, we applied concatenation and processed it with Gated Linear Units (GLUs) \cite{dauphin2017language}, in a novel way, to enrich features while maintaining the same number of parameters to avoid an increase in computation, as detailed in the subsection below.

\subsubsection{Integration of Basis and Spline Functions via Concatenation and Gated Linear Units (GLUs)}

In our Kolmogorov Arnold Network (KAN) layer implementation, we aim to enhance expressive power while maintaining the same feature dimensionality. To obtain this, we propose the following approach:

\begin{enumerate}
    \item \textbf{Concatenation}: Compute the outputs of $b(x)$ and $\text{spline}(x)$, then concatenate them:

    \[
    \text{concatenated\_output} = \text{concat}(w_b \cdot b(x), w_s \cdot \text{spline}(x))
    \]

    \item \textbf{Gated Linear Units (GLUs)}: Apply GLU to the concatenated output for non-linear transformation while maintaining the original feature dimensionality:

    \[
    \phi(x) = (W \cdot \text{concatenated\_output} + b) \odot \rho(V \cdot \text{concatenated\_output} + c)
    \]

    where $\odot$ denotes element-wise multiplication and $\rho$ represents the penalized tanh function with our customized implementation:
    
    \[
    \rho_{\alpha}(x) = \tanh(x) + \alpha \cdot \max(-x, 0)
    \]
    Here, \(\alpha\) is a hyperparameter that controls the penalty applied to the negative part of the input \(x\). It is worth mentioning that, while the sigmoid function is generally used in GLU, we integrated the penalized tanh as shown above, as it yielded better results during experiments.
\end{enumerate}

With these steps, we make different use of the residual connection mechanism, unlike the original KAN implementation, and implement the activation function \(\phi(x)\) as the concatenation of the basis function and the spline function. The construction ends with non-linear transformations to maintain the initial dimension of the concatenated inputs.

\begin{figure}[t]
  \centering
  \includegraphics[width=1\linewidth]{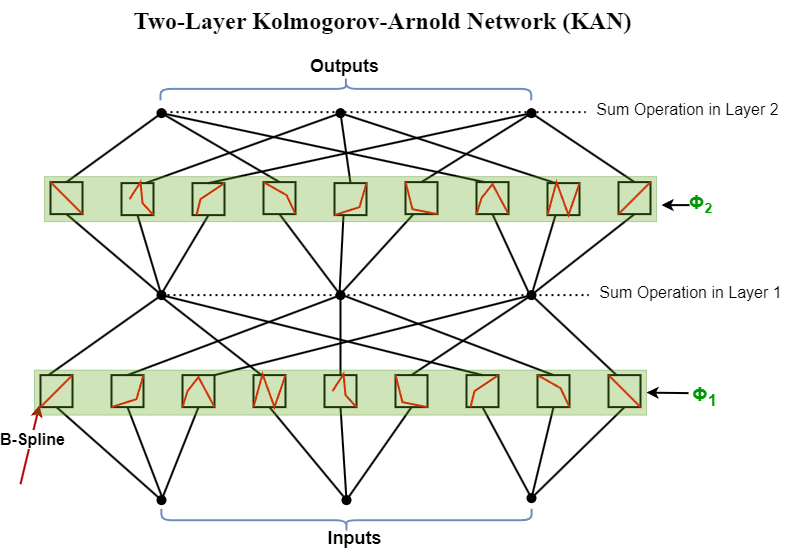}
  \caption{\textbf{Illustration of Two-Layer KAN.} $\Phi_1$ and $\Phi_2$ represent the collection of all 1D functions parametrized by B-Spline in layer 1 and layer 2, respectively.}
  \label{fig:TwoLayerKAN}
\end{figure}

Through these novel customizations of KAN, we finally construct a two-layer efficient KAN. Fig. \ref{fig:TwoLayerKAN} illustrates the process. Now, we will show how we integrate this obtained innovative two-layer efficient KAN into contrastive learning and GANs in unpaired image-to-image translation and give rise to the KAN-CUT model.

\subsection{Formulation of KAN-CUT Model}
Conforming to the principle of Generative Adversarial Networks (GANs), we integrate two neural network models: a generator and a discriminator. The generator's role is translating (generating) from input domain $A$ to output domain $B$, while the discriminator determines whether the provided image is genuine or artificial. Following recent literature in image-to-image translation \cite{zhu2017unpaired, park2020contrastive, han2021dual}, we have implemented ResNet-based encoder-decoder architecture of the generator.  

\begin{figure}[t]
  \centering
  \includegraphics[width=1\linewidth]{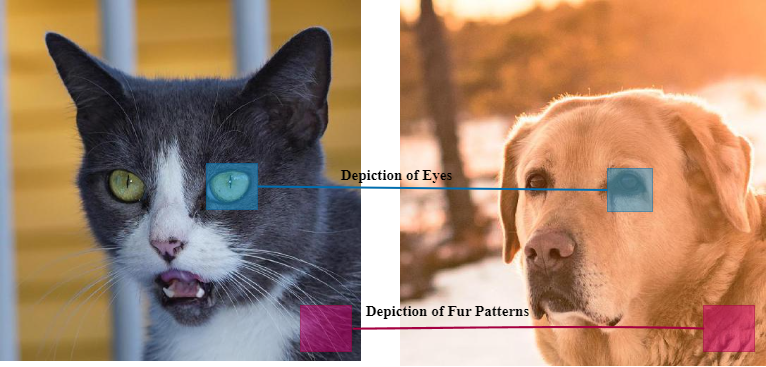}
  \caption{Illustration showing that images should share information based on patches. Images are sourced from the AFHQ dataset \cite{choi2020stargan} and are reproduced under the Creative Commons Attribution-NonCommercial 4.0 International License.}
  \label{fig:depictioninpatches}
\end{figure}

During the translation of an image $a$ \(in\) $A$ to an image $b$ \(in\) $B$, not only should the entire image share information, but the individual patches within the image should also share information while transforming domain-dependent information in the generated image. For instance, let's say we are translating an image from a cat to a dog. The generated dog image should share local information in the patches, meaning the eyes and nose should appear in the same locations as they were in the cat image (see Fig. \ref{fig:depictioninpatches}, where we depict the appearance of the eyes of a cat and a dog with blue-colored squares), while the obtained domain-dependent features such as fur patterns and facial structure should be unique to a dog (see Fig. \ref{fig:depictioninpatches}, where we highlight fur patterns with red-colored squares). We can achieve this effect of the images sharing patch-level information by maximizing information between corresponding patches while minimizing information in the non-corresponding patches. Since our work is based on an unsupervised mechanism, it is not straight forward to determine which patches are similar or corresponding and how to maximize the information. Thanks to the predictive mechanism of contrastive learning \cite{oord2018representation, chen2020simple}, we can use contrastive loss to successfully select the correct positive patch from a given set of patches in which there exists one positive and several negatives for a given query. More specifically, following \cite{park2020contrastive}, working in a vector representation, we construct an ($N$+1)-way classification problem, where the probability of positive being selected over negatives is achieved with cross-entropy loss given as:
\begin{equation}
L(q, p, \{n_i\}) = 
\\-\log 
\frac{\exp(q, p) / \tau)}{\exp(q, p) / \tau) + \sum_{i=1}^N \exp(q, n_i) / \tau)}
\end{equation}

where $q$, $p$ $\in \mathbb{R}^K$ and \(n_i\) $\in \mathbb{R}^{N \times K}$ are $K$-dimensional vectors corresponding to query, positive, and $N$ negatives, respectively. \( \tau \) is a temperature parameter to scale the distances between vectors. Following this procedure, we can select the correct query and positive vector from the set of vectors and maximize the mutual information between them in the embedding space. It is clear that we can obtain the selection and maximization of mutual information between vectors. Now, the question arises: how can we obtain the vectors or features where we can apply contrastive estimation?

\begin{figure*}[!htbp]
  \centering
  \includegraphics[width=0.85\linewidth]{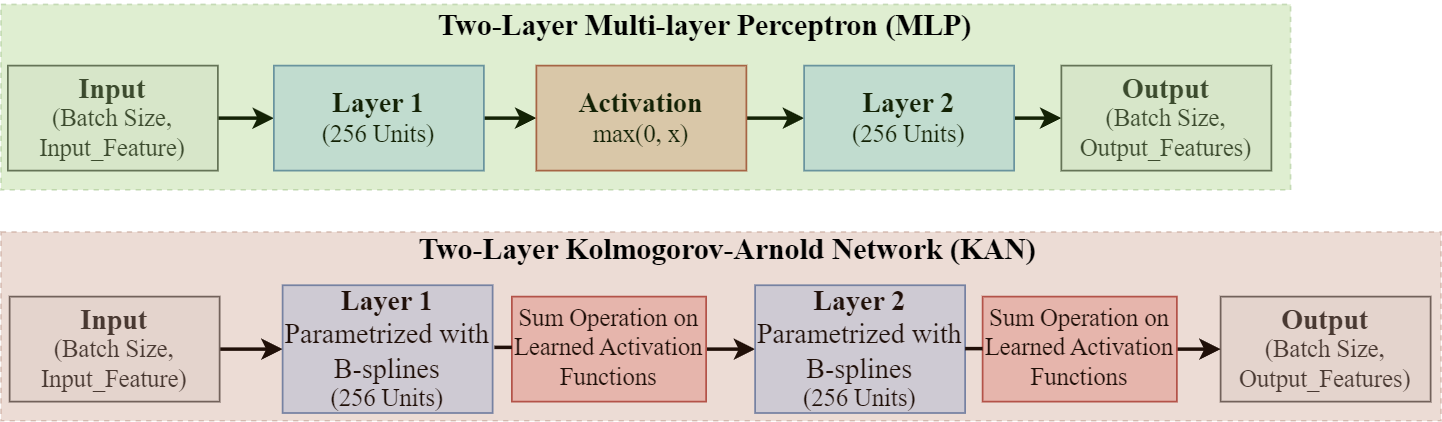}
  \caption{Comparison between the Two-Layer MLP and Two-Layer KAN architectures.}
  \label{fig:TwoLayerKANvsTwoLayerMLP}
\end{figure*}

\begin{figure*}[!htbp]
  \centering
  \includegraphics[width=0.65\linewidth]{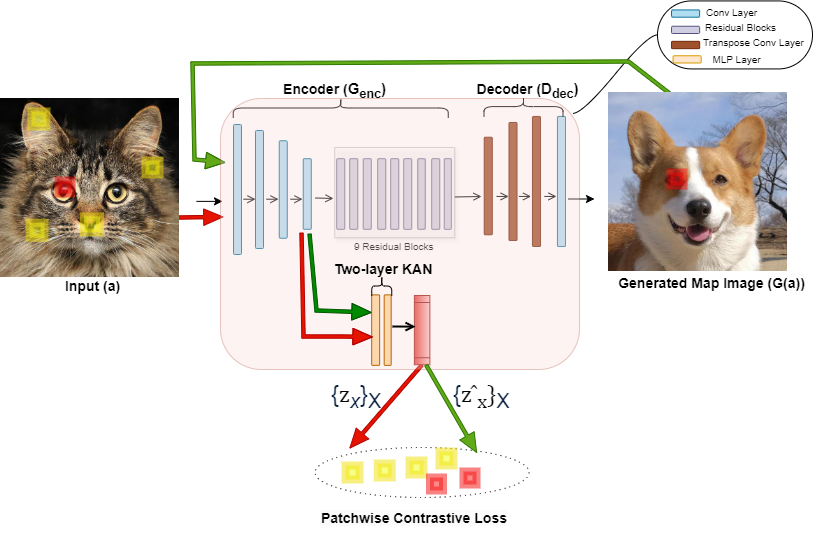}
  \caption{Architecture of Generator used in KAN-CUT. The cat and dog images are sourced from the AFHQ dataset \cite{choi2020stargan} and are reproduced under the Creative Commons Attribution-NonCommercial 4.0 International License.}
  \label{fig:Generator}
\end{figure*}

As mentioned in \cite{park2020contrastive}, since images are processed in the generator $G$, we can utilize features from intermediate layers of the generator's encoder, $G_{\text{enc}}$. Inspired by \cite{park2020contrastive} and \cite{chen2020simple}, we select $X$ layers from the encoder and pass them to our innovative two-layer efficient KAN instead of a two-layer MLP (see Fig. \ref{fig:TwoLayerKANvsTwoLayerMLP} illustrating the difference in layer implementation between KAN and MLP), producing a more informative stack of features. For instance, let's say we feed image $a$ to the generator $G$ and generate image $G(a)$. To obtain stacks of features from which the positive feature or vector is chosen, we select the $x$-th layer from the encoder through which $a$ is being processed and pass it to the two-layer KAN network $K_l$ to produce a stack of features $\{z_x\}_X = \{K_x(G_{\text{enc}}^{l}(a))\}_X$ (as depicted with the red-colored arrow in Fig. \ref{fig:Generator}).

In each layer, there are several spatial locations quantified by $S_x$, where $s \in \{1, \ldots, S_x\}$. We refer to the positive feature as $z_x^s \in \mathbb{R}^{C_x}$ and the negative features as $z_x^{S \setminus s} \in \mathbb{R}^{(S_x-1) \times C_x}$, where $C_x$ denotes the channel count at each layer. To denote a query vector or feature in this instance, we feed the generated image $G(a)$ to the same encoder, select the $x$-th layer, and pass it to the two-layer KAN network $K_l$ to produce another stack of features, $\{\hat{z}_x\}_X = \{K_x(G_{\text{enc}}^{l}(G(a)))\}_X$ (as depicted with green-colored arrow in Fig. \ref{fig:Generator}). From this feature stack, we refer to a corresponding query point as $\hat{z}_x^s$. It is worth mentioning that negatives vectors are drawn from the same input image, following \cite{park2020contrastive}. Once we have the query, positive, and negative vectors, we can perform constrastive learning using the loss function given above in equation (1). The above procedure is performed for an instance of a spatial location and from a layer; however, we need to do this for all spatial locations and layers. For that, we use PatchNCE loss as presented in \cite{park2020contrastive}, which can be expressed as:
\begin{equation}
%\begin{aligned}
L_{\text{PatchNCE}}(G, K, A) = \mathbb{E}_{a \sim A} \Bigg[ \sum_{x=1}^{X} \sum_{s=1}^{S_x} X\Big(\hat{z}_{x}^{s}, z_{x}^{s}, z_{x}^{S \setminus s}\Big) \Bigg]
%\end{aligned}
\end{equation}

\subsubsection{Adversarial Loss for Generation of Realistic-Looking Images}
We can achieve mutual information maximization at the patch level with the above procedure using contrastive learning and the two-layer efficient KAN. For high-quality image generation, we use adversarial loss from generative adversarial networks (GANs). The initial work in GANs was introduced in \cite{goodfellow2014generative}, and this architecture is considered as Vanilla GAN. The authors introduced the use of a sigmoid cross-entropy loss for GANs. This loss can sometimes cause vanishing gradients when data samples fall within the correct classification boundary but remain distant from the actual data distribution. To mitigate this problem, \cite{mao2017least} proposed Least Squares Generative Adversarial Networks (LSGANs), replacing the binary cross-entropy loss with a least squares loss. Following principles of \cite{goodfellow2014generative} and \cite{mao2017least}, to guarantee the generation of authentic images perceived by humans in domain $B'$, such that $B' \approx B$ for the given input images from domain $A$, we utilize an adversarial loss based on the least squares loss. The adversarial loss, also known as LSGAN loss, comprises two main components: the loss for the generator and the loss for the discriminator. These losses guide the backpropagation process through which the neural networks are trained and updated. The general procedure involves training the generator to generate images in the target domain that are identical to real images, whereas the discriminator learns to differentiate real images from generated ones.

The adversarial loss can be formulated as:
\begin{equation}
\begin{aligned}
L_{\text{LSGAN}}(D) &= \frac{1}{2} \mathbb{E}_{b \sim B}[(D(b) - 1)^2] + \frac{1}{2} \mathbb{E}_{a \sim A}[(D(G(a)))^2] \\
L_{\text{LSGAN}}(G) &= \frac{1}{2} \mathbb{E}_{a \sim A}[(D(G(a)) - 1)^2]
\end{aligned}
\end{equation}

In these equations, \( D(b) \) represents the discriminator's decision for a ground truth image \( b \) from the target domain, where the label for real images is set to 1. \( G(a) \) is the generated image from the input image \( a \) from the input domain, where the label for fake images is set to 0. The generator \( G \) aims to minimize \( L_{\text{LSGAN}}(G) \), making the discriminator believe that the generated images are real by pushing \( D(G(a)) \) towards 1. The discriminator \( D \) aims to minimize \( L_{\text{LSGAN}}(D) \), correctly distinguishing real images from generated ones by pushing \( D(b) \) towards 1 and \( D(G(a)) \) towards 0. This results in a minimax game between the two, similar to the initial work in \cite{goodfellow2014generative}, promoting the generation of high-quality, realistic images.

\subsubsection{Final Loss Functions}
Following the same principle in \cite{park2020contrastive}, we utilize three different loss functions involving only one generator and discriminator. These loss functions include the adversarial loss for generating realistic-looking images, the PatchNCE loss, \(L_{\text{PatchNCE}}\)(G, K, A), for ensuring patch-level correspondence, and a similar PatchNCE loss, \(L_{\text{PatchNCE}}\)(G, K, B), to prevent inappropriate translation by the generator, similar to the identity loss presented in \cite{zhu2017unpaired}. The combined final loss function is formulated as follows:
\begin{equation}
\begin{aligned}
L_{\text{final}} = & L_{\text{LSGAN}}(G, D, A, B) + \\
                  & \lambda_{\text{PatchNCE-A}} L_{\text{PatchNCE}}(G, K, A) + \\
                  & \lambda_{\text{PatchNCE-B}} L_{\text{PatchNCE}}(G, K, B)
\end{aligned}
\end{equation}

\section{Experiments}
For evaluation, we selected two well-known, widely utilized, and publicly accessible datasets for research in image-to-image translation: \(Horse\rightarrow Zebra\) and \(Cat\rightarrow Dog\).

\subsection{Datasets}
\paragraph{\(Horse\)\textrightarrow{}\(Zebra\)}
First introduced in CycleGAN \cite{zhu2017unpaired}, this dataset comprises 1067 horse images and 1344 zebra images, resulting in 2403 training images. Additionally, there are 120 horse images and 140 zebra images, totaling 260 test images.

\paragraph{\(Cat\)\textrightarrow{}\(Dog\)}
First introduced in StarGAN V2 \cite{choi2020stargan}, this dataset is a subset of AFHQ and includes 5000 training images and 500 test images for each category.

\subsection{Evaluation Metrics}
For the evaluation metric, we selected the Fréchet Inception Distance (FID) \cite{heusel2017gans} score. It is one of the most utilized metrics for assessing the quality of images produced by GANs. The FID score evaluates the distance between feature vectors calculated for real and fake (generated) images, utilizing a pre-trained Inception v3 \cite{szegedy2016rethinking} network. A lower FID score signifies higher similarity to the real images, thus representing better quality of the generated images.

\subsection{Experimental Environment and Baselines}

\paragraph{Experiment Setting}
Our experiments were performed in the Python 3.6.8 environment using the PyTorch framework for all facets of training and testing. The computational tasks were carried out on a system comprising NVIDIA A100-PCI GPUs, each featuring 80 GB of HBM2 memory. The computation tasks were facilitated using CUDA version 12.3 and NVIDIA driver version 545.23.08. All the models were trained for 400 epochs using the Adam optimization algorithm \cite{kingma2014adam}, set at a learning rate of 0.0001. During the training of our model, KAN-CUT, we chose \(\lambda_{\text{PatchNCE-A}} = 1\) and \(\lambda_{\text{PatchNCE-B}} = 1\).
 
\paragraph{Baselines}
For our comparative analysis, we selected well-known GAN models as our baselines: CycleGAN \cite{zhu2017unpaired}, MUNIT \cite{huang2018multimodal}, SelfDistance \cite{benaim2017one}, GCGAN \cite{fu2019geometry}, CUT \cite{park2020contrastive}, and DCLGAN \cite{han2021dual}. It is worth noting that all models were trained in our setup except for MUNIT \cite{huang2018multimodal}, SelfDistance \cite{benaim2017one}, and GCGAN \cite{fu2019geometry}, for which the FID scores were recorded from \cite{han2021dual}, where DCLGAN was proposed.

\subsection{Results}
The quantitative results from each baseline model, including the proposed KAN-CUT, are presented in Table \ref{tab:Horse2Zebra_Cat2Dog}. It can be seen that KAN-CUT outperforms all the models on both selected datasets, with an FID score of 40.2 on the \(Horse\)\textrightarrow{}\(Zebra\) dataset, and 59.55 on the \(Cat\)\textrightarrow{}\(Dog\) dataset.

\begin{figure*}[!t]
    \centering
    \begin{tabular}{ccccc}
        % Header row
        \textbf{Input} & \textbf{CycleGAN} & \textbf{CUT} & \textbf{DCLGAN} & \textbf{KAN-CUT} \\
        
        % First row of images
        \includegraphics[width=0.12\textwidth]{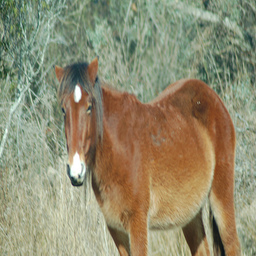} &
        \includegraphics[width=0.12\textwidth]{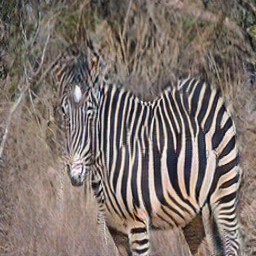} &
        \includegraphics[width=0.12\textwidth]{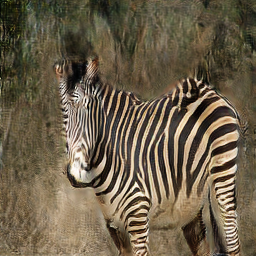} &
        \includegraphics[width=0.12\textwidth]{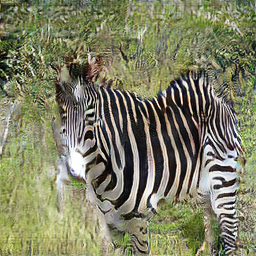} &
        \includegraphics[width=0.12\textwidth]{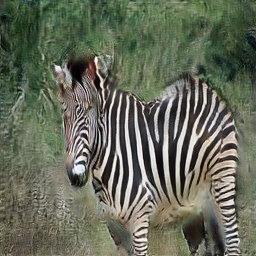} \\
        
        % Second row of images
        \includegraphics[width=0.12\textwidth]{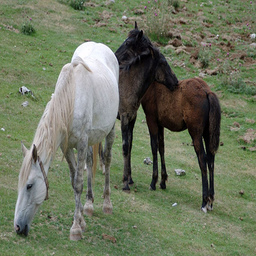} &
        \includegraphics[width=0.12\textwidth]{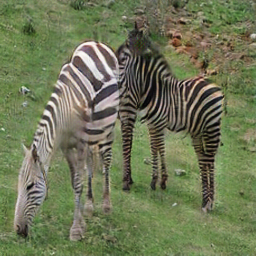} &
        \includegraphics[width=0.12\textwidth]{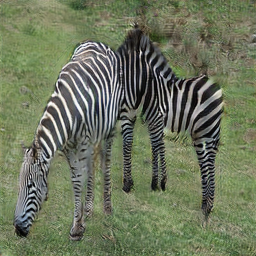} &
        \includegraphics[width=0.12\textwidth]{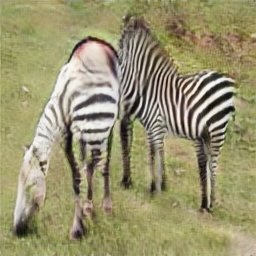} &
        \includegraphics[width=0.12\textwidth]{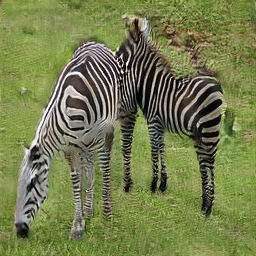} \\
        
        % Third row of images
        \includegraphics[width=0.12\textwidth]{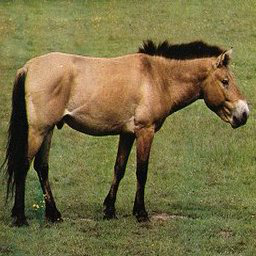} &
        \includegraphics[width=0.12\textwidth]{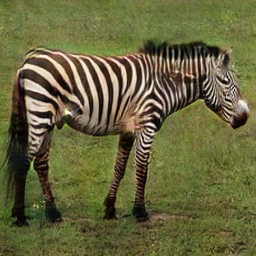} &
        \includegraphics[width=0.12\textwidth]{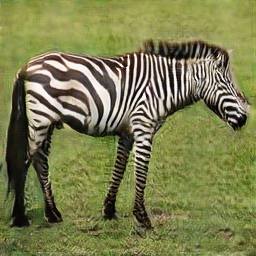} &
        \includegraphics[width=0.12\textwidth]{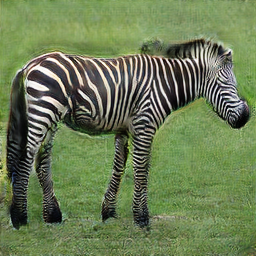} &
        \includegraphics[width=0.12\textwidth]{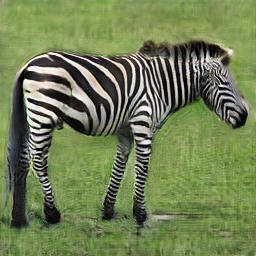} \\
        
        % Fourth row of images
        \includegraphics[width=0.12\textwidth]{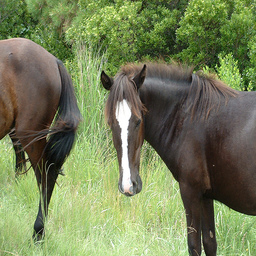} &
        \includegraphics[width=0.12\textwidth]{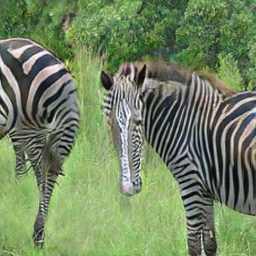} &
        \includegraphics[width=0.12\textwidth]{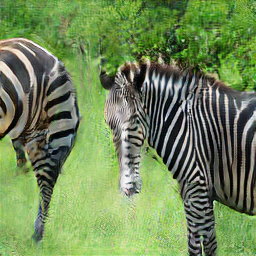} &
        \includegraphics[width=0.12\textwidth]{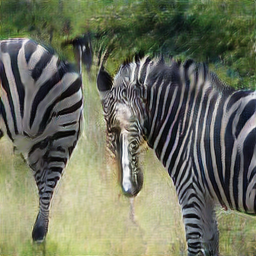} &
        \includegraphics[width=0.12\textwidth]{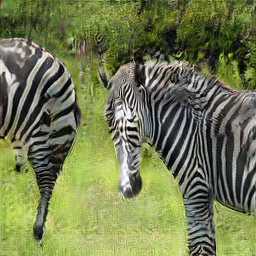} \\

        %Images of Cat2Dog
          % Fifth row of images
        \includegraphics[width=0.12\textwidth]{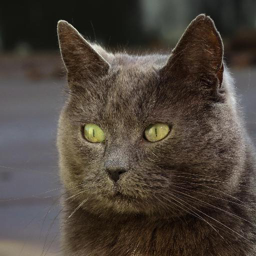} &
        \includegraphics[width=0.12\textwidth]{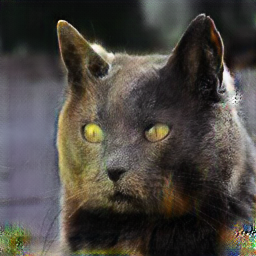} &
        \includegraphics[width=0.12\textwidth]{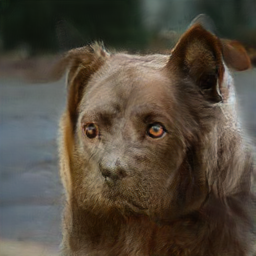} &
        \includegraphics[width=0.12\textwidth]{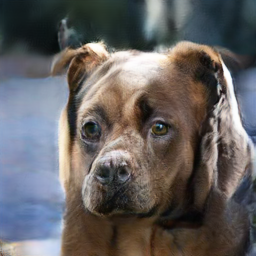} &
        \includegraphics[width=0.12\textwidth]{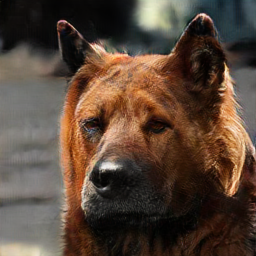} \\
        
        % Third row of images
        \includegraphics[width=0.12\textwidth]{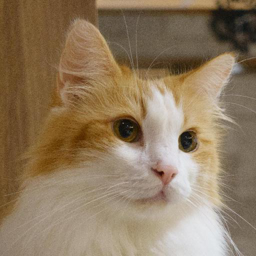} &
        \includegraphics[width=0.12\textwidth]{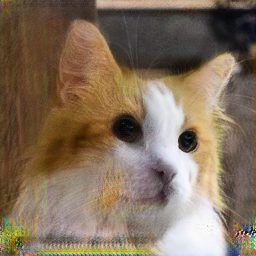} &
        \includegraphics[width=0.12\textwidth]{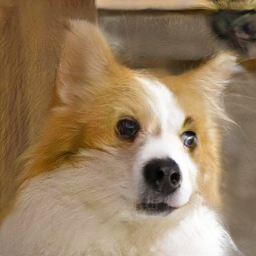} &
        \includegraphics[width=0.12\textwidth]{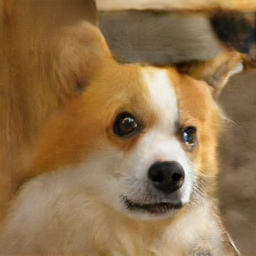} &
        \includegraphics[width=0.12\textwidth]{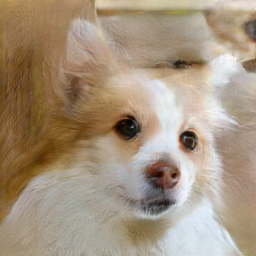} \\
        
        % Fourth row of images
        \includegraphics[width=0.12\textwidth]{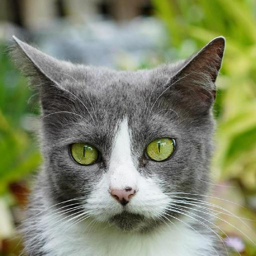} &
        \includegraphics[width=0.12\textwidth]{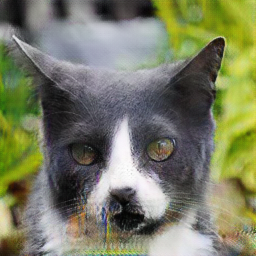} &
        \includegraphics[width=0.12\textwidth]{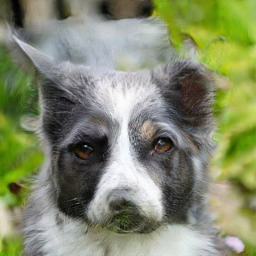} &
        \includegraphics[width=0.12\textwidth]{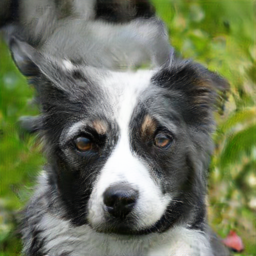} &
        \includegraphics[width=0.12\textwidth]{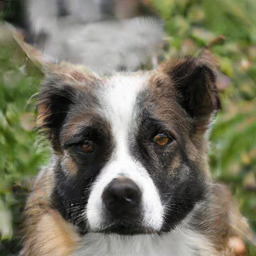} \\
        
        % Fifth row of images
        \includegraphics[width=0.12\textwidth]{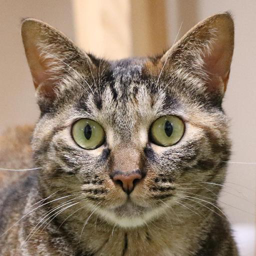} &
        \includegraphics[width=0.12\textwidth]{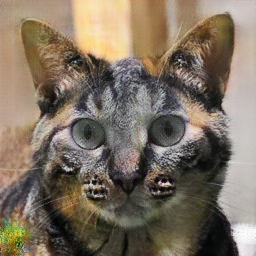} &
        \includegraphics[width=0.12\textwidth]{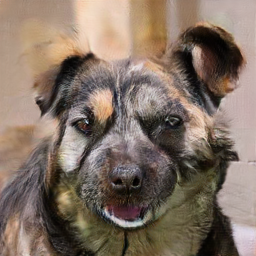} &
        \includegraphics[width=0.12\textwidth]{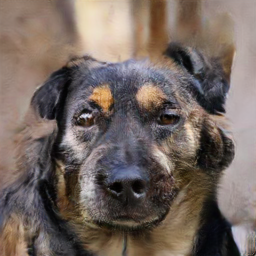} &
        \includegraphics[width=0.12\textwidth]{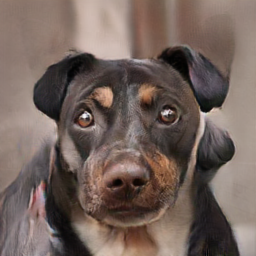} \\
    \end{tabular}
    \caption{\textbf{Comparative Results of \(\mathbf{Horse\rightarrow Zebra}\) and \(\mathbf{Cat\rightarrow Dog}\)}. The figure presents a side-by-side comparison of zebra images generated by various models, including KAN-CUT, in the first four rows, and the same comparison of dog images generated by those models in the fifth to eighth rows. The figure highlights the effectiveness of each approach in synthesizing accurate details such as structure and color. Images from the \(Horse\rightarrow Zebra\) dataset are sourced from ImageNet \cite{5206848} and are reproduced under the ImageNet terms of access, which allows use for non-commercial research and educational purposes. Images from the \(Cat\rightarrow Dog\) dataset are sourced from the AFHQ dataset \cite{choi2020stargan} and are reproduced under the Creative Commons Attribution-NonCommercial 4.0 International License.}
    \label{fig:image-comparison}
\end{figure*}

Additionally, our study presents a visualization of the performance of each model in generating zebra images from horse images and dog images from cat images, as depicted in Fig. \ref{fig:image-comparison}. As mentioned above, we did not run the experiments for MUNIT \cite{huang2018multimodal}, SelfDistance \cite{benaim2017one}, and GCGAN \cite{fu2019geometry} in our work and could not find the resulting images generated by these models online. Therefore, the visualization comparison was done against CycleGAN, DCLGAN, and CUT. The KAN-CUT model is able to generate higher quality zebra images based on stripes, structure, and color, compared to all these selected baseline models. 

\begin{table}
  \centering
  {\small{
  \begin{tabular}{@{}lcc@{}}
    \toprule
    Method & Horse$\rightarrow$Zebra & Cat$\rightarrow$Dog \\
           & FID $\downarrow$ & FID $\downarrow$ \\
    \midrule
    CycleGAN \cite{zhu2017unpaired}& 66.8  & 85.9 \\
    MUNIT \cite{huang2018multimodal}                  & 133.8 & 104.4\\
    SelfDistance \cite{benaim2017one}           & 80.8  & 144.4\\
    GCGAN \cite{fu2019geometry}    & 86.7  & 96.6 \\
    CUT \cite{park2020contrastive} & 45.5  & 76.2 \\
    DCLGAN \cite{han2021dual}      & 43.2  & 60.7 \\
    KAN-CUT (Ours)                 & 40.2  & 59.55 \\
    \bottomrule
  \end{tabular}
  }}
  \caption{Comparison of the performance of different GAN models with the proposed model, KAN-CUT, on the Horse$\rightarrow$Zebra and Cat$\rightarrow$Dog datasets. The performance is measured using the FID score (lower is better).}
  \label{tab:Horse2Zebra_Cat2Dog}
\end{table}

\section{Conclusion, Limitation, and Future Work}
GANs, being among the most prominent generative models, have played a major role in the success of Image-to-Image (I2I) generation (translation), a subdomain of Generative AI. Despite their success, there are several areas requiring further research, particularly in terms of accuracy, computational efficiency, and knowledge transferability. Focusing primarily on accuracy and demonstrating the efficacy of Kolmogorov-Arnold Networks (KAN) \cite{liu2024kan} in the image-to-image translation domain, we propose the KAN-CUT model. This novel model replaces the two-layer MLP in the CUT model \cite{park2020contrastive} with a two-layer, efficient, and customized KAN.

As shown in the Results section, our proposed KAN-CUT model outperforms all selected baseline GAN models on two well-known research datasets, \(Horse\)\textrightarrow{}\(Zebra\) and \(Cat\)\textrightarrow{}\(Dog\), in both quantitative and qualitative assessments. Notably, KAN-CUT, being an upgraded version of CUT, achieved a superior FID score by 5.3 points on the \(Horse\)\textrightarrow{}\(Zebra\) dataset and by 16.65 points on the \(Cat\)\textrightarrow{}\(Dog\) dataset. To the best of our knowledge, our study is the first to propose and demonstrate the applicability of KAN in image-to-image translation, potentially paving the way for numerous applications of KAN in Generative AI, which we find very promising.

While the present study has been successful, it has some limitations. Our primary goal was to demonstrate the applicability of KAN \cite{liu2024kan} and to showcase its improved performance compared to MLP. We do not claim improvement over the state-of-the-art performance in the image-to-image translation domain. Additionally, the KAN-CUT model's performance has been demonstrated on only two datasets, \(Horse\)\textrightarrow{}\(Zebra\) and \(Cat\)\textrightarrow{}\(Dog\), due to time and resource constraints. Experimental validation on other datasets remains an area for future work.

\section*{Acknowledgement}
This material is based in part upon work supported by the National Science Foundation under Grant Nos. CNS-2018611 and CNS-1920182.

\nocite{*}
\bibliographystyle{IEEEtran}
%\bibliography{bibliography} 
\input{main.bbl}

\end{document}

%% file: main.bbl
% Generated by IEEEtran.bst, version: 1.14 (2015/08/26)